\documentclass[conference]{IEEEtran}
\IEEEoverridecommandlockouts

\usepackage{cite}
\usepackage{amsmath,amssymb,amsfonts}
\usepackage{graphicx}
\usepackage{textcomp}
\usepackage{xcolor}

\usepackage{multirow}
\usepackage{graphicx}
\usepackage[normalem]{ulem}
\useunder{\uline}{\ul}{}
\usepackage{xcolor}
\usepackage{subfig}
\usepackage{booktabs,arydshln}
\usepackage{xurl}
\usepackage{enumitem}
\usepackage{algorithm}
\usepackage[noend]{algpseudocode}

\begin{document}

\title{Node Importance Estimation Leveraging LLMs for Semantic Augmentation in Knowledge Graphs\\

\thanks{*Corresponding author.}
}

\author{\IEEEauthorblockN{1\textsuperscript{st} Xinyu Lin}
\IEEEauthorblockA{\textit{School of Computing and Artificial Intelligence} \\
\textit{Fuyao University of Science and Technology}\\
Fuzhou, China \\
xinyu.lin99cn@gmail.com}
\and
\IEEEauthorblockN{1\textsuperscript{st} Tianyu Zhang}
\IEEEauthorblockA{\textit{Department of Automation} \\
\textit{Tsinghua University}\\
Beijing, China \\
zhangty2016@gmail.com}
\and
\IEEEauthorblockN{3\textsuperscript{rd} Chengbin Hou*}
\IEEEauthorblockA{\textit{School of Computing and Artificial Intelligence} \\
\textit{Fuyao University of Science and Technology}\\
Fuzhou, China \\
chengbin.hou10@foxmail.com}
\and
\IEEEauthorblockN{4\textsuperscript{th} Jinbao Wang}
\IEEEauthorblockA{\textit{National Engineering Laboratory} \\
\textit{for Big Data System Computing Technology}\\
\textit{Shenzhen University}, China\\
wangjb@szu.edu.cn
}
\and
\IEEEauthorblockN{5\textsuperscript{th} Jianye Xue}
\IEEEauthorblockA{\textit{Department of Automation} \\
\textit{Tsinghua University}\\
Beijing, China \\
xjy23@mails.tsinghua.edu.cn
}
\and
\IEEEauthorblockN{6\textsuperscript{th} Hairong Lv}
\IEEEauthorblockA{\textit{Department of Automation} \\
\textit{Tsinghua University}\\
Beijing, China \\
lvhairong@tsinghua.edu.cn}
}

\maketitle

\begin{abstract}
Node Importance Estimation (NIE) is a task that quantifies the importance of node in a graph. Recent research has investigated to exploit various information from Knowledge Graphs (KGs) to estimate node importance scores. However, the semantic information in KGs could be insufficient, missing, and inaccurate, which would limit the performance of existing NIE models. To address these issues, we leverage Large Language Models (LLMs) for semantic augmentation thanks to the LLMs' extra knowledge and ability of integrating knowledge from both LLMs and KGs. To this end, we propose the LLMs Empowered Node Importance Estimation (LENIE) method to enhance the semantic information in KGs for better supporting NIE tasks. To our best knowledge, this is the first work incorporating LLMs into NIE. Specifically, LENIE employs a novel clustering-based triplet sampling strategy to extract diverse knowledge of a node sampled from the given KG. After that, LENIE adopts the node-specific adaptive prompts to integrate the sampled triplets and the original node descriptions, which are then fed into LLMs for generating richer and more precise augmented node descriptions. These augmented descriptions finally initialize node embeddings for boosting the downstream NIE model performance. Extensive experiments demonstrate LENIE’s effectiveness in addressing semantic deficiencies in KGs, enabling more informative semantic augmentation and enhancing existing NIE models to achieve the state-of-the-art performance. The source code of LENIE is freely available at \url{https://github.com/XinyuLin-FZ/LENIE}.
\end{abstract}


\section{Introduction}
Node Importance Estimation (NIE) aims to evaluate the significance of nodes within a graph, providing a crucial foundation for practical applications such as resource allocation, data management, and recommendation systems \cite{liu2021survey, park2019estimating, huang2022traffic}. Earlier methods for the NIE task, e.g., Google's PageRank \cite{page1999pagerank}, compute node importance scores by analyzing the quantity and quality of incoming edges to help users identify the most relevant pages. However, these early NIE methods, including random walk-based \cite{page1999pagerank, haveliwala2002topic, tong2008random} and centrality-based \cite{nieminen1974centrality, sabidussi1966centrality, das2018study, saxena2020centrality} approaches, are primarily designed for homogeneous graph data and evaluate node importance solely based on graph topology. With the growing complexity of application scenarios, heterogeneous graphs, which contain richer information than homogeneous graphs, have been widely utilized \cite{zou2020survey}. Effectively leveraging richer information in graph data to more comprehensively and accurately assess node importance has become a key focus of recent NIE research.

Knowledge Graphs (KGs), a prominent type of complex heterogeneous graphs, encompass real-world knowledge through triplets (head node, relational edge, tail node), which contain rich structural and semantic information. Many recent studies have focused on leveraging this wealth of information in KGs to enhance NIE tasks \cite{park2019estimating, park2020multiimport, huang2022estimating, liu2023node, huang2021representation, zhang2024label, chen2024deep}. GENI \cite{park2019estimating}, the first NIE method for KGs, employs a GNN-based aggregation and update mechanism to effectively capture structural information within KGs, achieving promising NIE results. Later on, other research efforts have explored incorporating more information from KGs, such as semantic content \cite{huang2021representation}, multiple input signals \cite{park2020multiimport}, local and global features \cite{huang2022estimating}, and attention to the most important nodes \cite{zhang2024label}, to enhance NIE performance. Notably, several studies have demonstrated that exploiting node descriptions can improve the model performance of NIE \cite{huang2021representation, huang2022estimating, chen2024deep, zhang2024label}, further verifying the benefits of integrating semantic information from KGs into NIE tasks.

\begin{figure}
    \centering
    \includegraphics[scale=0.35]{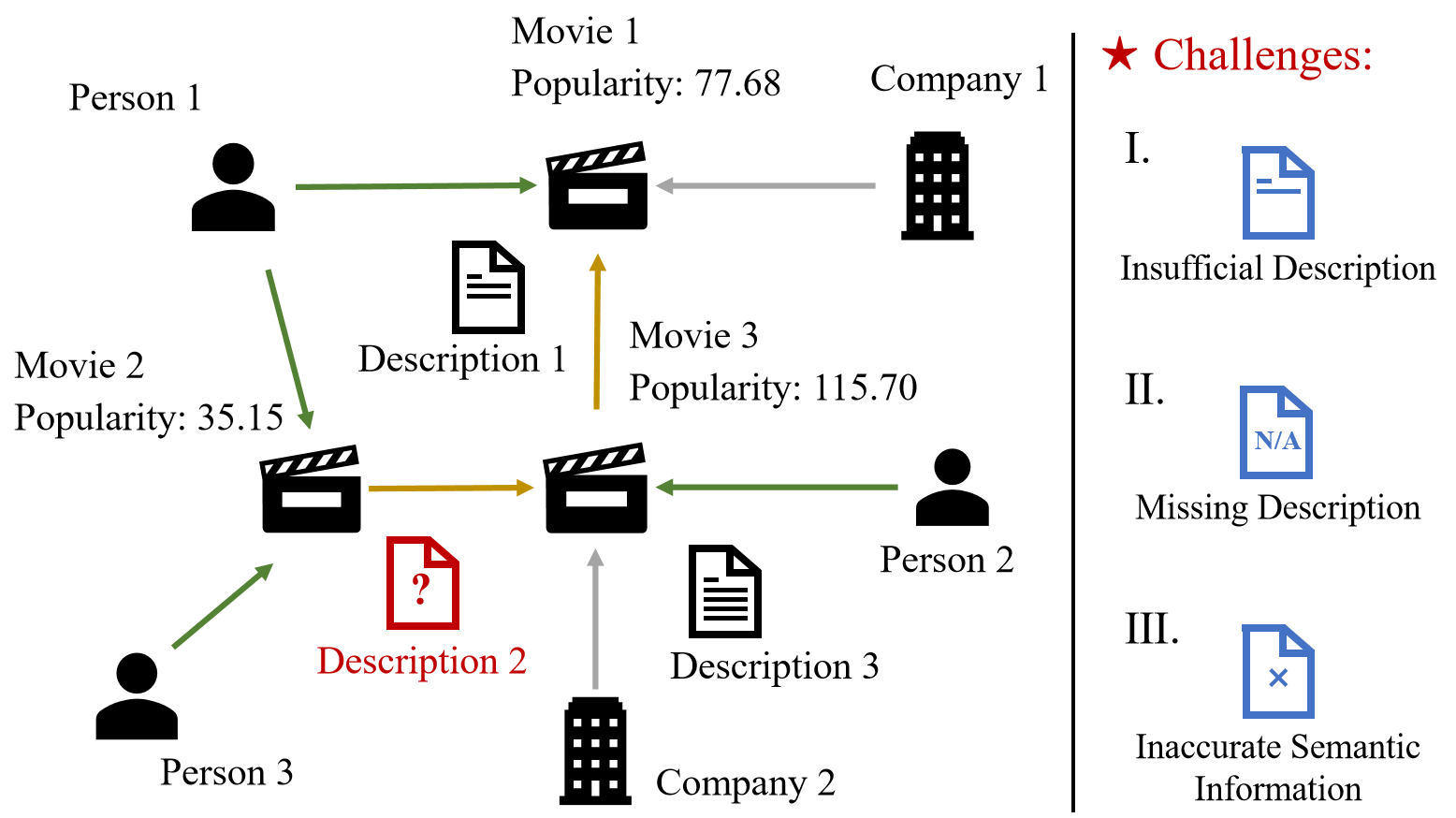}
    \caption{A small movie KG with different types of edges (color-coded) and nodes. Popularity indicates the importance score for movie nodes, each with a description, but the description (marked in red) may be insufficient, missing, or inaccurate.}
    \label{Fig:1}
\end{figure}

However, KG data is often incomplete \cite{jain2020domain, xue2022knowledge}, leading to deficiencies in their semantic information. As illustrated in Figure \ref{Fig:1}, the semantic information of nodes in KGs would encounter issues such as insufficient descriptions, missing descriptions, and inaccuracies. Specifically, the insufficient descriptions refer to short or incomplete textual descriptions of nodes in KGs; the missing descriptions indicate the absence of textual descriptions for some nodes; and the inaccuracies refer to inaccurate semantic information of nodes regarding their description texts or related triplets. These issues would limit the performance improvement of existing NIE methods that make use of semantic information from KGs.

Motivated by the recent advancements in Large Language Models (LLMs), we intend to adopt LLMs to improve KGs by addressing
the incompleteness and inaccuracies \cite{pan2024unifying,paulheim2017knowledge}. LLMs excel at providing extra knowledge beyond the given KG, since they are trained on vast amounts of texts \cite{achiam2023gpt,touvron2023llama}. Meanwhile, the techniques like in-context learning and retrial augmented generation \cite{dong2022survey,lewis2020retrieval} enable LLMs to smoothly integrate with KGs, thereby assisting LLMs to generate more precise and specific contents for the given KG. These abilities make LLMs possible to tackle the issues of insufficient descriptions, missing descriptions, and inaccurate semantic information in KGs. Consequently, we introduce LLMs to alleviate these issues restricting the performance of NIE models. 

To enrich the semantic information for a given KG and accordingly further support the NIE task, we propose the \underline{L}LMs \underline{E}mpowered \underline{N}ode \underline{I}mportance \underline{E}stimation (LENIE) method. Concretely, we first develop a clustering-based triplet sampling strategy to extract the semantic knowledge of a node from the KG, which serves as the contexts surrounding that node for the given KG; the clustering-based sampling strategy is used to keep the diversity and coverage of the knowledge offered by the given KG. Second, the node-specific adaptive prompts are designed to combine the sampled triplets with the original node descriptions, enabling LLMs to generate richer and more precise augmented node descriptions, thus utilizing LLMs' understanding of real-world entity and the ability of integrating domain knowledge sampled from the given KG. Finally, the augmented descriptions are adopted to initialize node embeddings with richer and more precise information to boost the performance of various existing NIE models.

The main contributions of this work are as follows:
\begin{itemize}
 \item Existing NIE methods have not considered the situation where KG data is incomplete, as evidenced by the fact that the semantic information is possibly insufficient, missing, or inaccurate. We accordingly introduce a novel framework that leverages LLMs for semantic augmentation in KGs for node importance estimation.
 \item Technically, we develop a clustering-based triplet sampling strategy to effectively keep the diversity and coverage of the knowledge offered by the given KG. We also design the node-specific adaptive prompts to smoothly integrate the sampled triplets with the original node descriptions, aiming to guide LLMs in generating richer and more precise augmented node descriptions for further improving the performance of NIE models.
 \item Extensive experiments demonstrate that LENIE can boost the performance of existing NIE models across various KGs and metrics, as well as achieve the new state-of-the-art NIE performance. Furthermore, we also conduct experiments to confirm the effectiveness of key designs in LENIE and show the capability of LENIE in addressing the semantic deficiencies in KGs.
 \item To our best knowledge, this is the first attempt to incorporate LLMs into the NIE task. And to facilitate future research in NIE and LLM communities, we release our source code at \url{https://github.com/XinyuLin-FZ/LENIE}.
 \end{itemize}

\section{Related Work}\label{related work}

\subsection{Node Importance Estimation}

NIE has evolved from applications in simple homogeneous graphs to complex heterogeneous graphs, with a growing focus on leveraging the wealth of information within graphs to more accurately assess node importance.
Earlier work primarily focused on homogeneous graphs \cite{page1999pagerank, haveliwala2002topic, tong2008random, nieminen1974centrality, sabidussi1966centrality}, where nodes and edges each have a single type, and these methods are typically categorized into random walk-based and centrality-based methods. PageRank (PR) \cite{page1999pagerank} is a classic random walk-based method that captures the structural information of a graph and evaluates node importance through random walks with equal probability on the graph. Personalized PageRank (PPR) \cite{haveliwala2002topic}  extends the standard random walk approach by incorporating specific topic information. Random Walk with Restart \cite{tong2008random} enhances PR by adding a restart mechanism. Unlike random walk-based methods, centrality-based methods tend to identify critical nodes in the graph. Degree centrality \cite{nieminen1974centrality} evaluate a node's importance by its number of direct connections. Closeness centrality \cite{sabidussi1966centrality} measures a node's importance by the inverse of the sum of its shortest path distances to all other nodes in the network. More concepts and improved versions of centrality-based methods can be found in \cite{das2018study, saxena2020centrality}. In summary, both methods mentioned above can assess node importance in graphs. However, they focus on homogeneous graphs, as they rely on the topology of the graph structure for the NIE task, making it challenging to leverage the richer information available in more complex graphs.

In recent years, GNN-based models have excelled in handling complex heterogeneous graphs like KGs by aggregating and propagating rich information within the graph, achieving state-of-the-art results in numerous NIE tasks \cite{park2019estimating, park2020multiimport, huang2022estimating, liu2023node, huang2021representation, zhang2024label, chen2024deep}.
GENI \cite{park2019estimating} is the first to apply GNN for estimating node importance in KGs, utilizing an attention mechanism to aggregate graph structure information for the NIE task.
MultiImport \cite{park2020multiimport} proposes a GNN-based model that learns from multiple input signals to infer node importance in KGs. HIVEN \cite{huang2022estimating} considers the importance of nodes as heterogeneous values and evaluates them by aggregating both local and global information. MCRL \cite{liu2023node} obtains node representations from multiple perspectives and infers node importance by combining contrastive learning techniques. Additionally, many NIE works aim to leverage the rich semantic information in KGs by encoding the description text of nodes into their embeddings, thereby assisting downstream NIE models in achieving better performance \cite{chen2024deep, huang2022estimating, huang2021representation, zhang2024label}. RGTN \cite{huang2021representation} specifically designed a separate encoder to capture the semantic information, i.e., descriptions of nodes, and combines it with structural information to evaluate node importance in KGs. LICAP \cite{zhang2024label} builds upon RGTN by generating contrastive samples using continuous labels and further pre-training the semantic embeddings of nodes in KGs, achieving state-of-the-art performance in both regression and ranking metrics for NIE tasks. Thus, effectively utilizing the information in KGs, particularly the semantic information, is advantageous for improving NIE models.

Previous NIE methods rely on node description texts from datasets for semantic information, which often suffer from being insufficient, missing, or inaccurate. Unlike previous NIE works, our study focuses on enriching the semantic information in KGs. Specifically, we incorporate the context of KGs into LLMs through effective clustering-based triplet sampling strategy to generate more accurate and comprehensive descriptions of nodes, thereby enhancing the performance of downstream NIE models. To the best of our knowledge, we are the first to integrate LLMs with the NIE task, providing a novel approach to estimating node importance in KGs.

\subsection{Large Language Models for Knowledge Graphs}

LLMs, trained on vast amounts of textual data, exhibit strong capabilities in both text comprehension and generation, achieving outstanding performance across a wide range of tasks in diverse fields \cite{min2023recent, chang2024survey, raiaan2024review}. The popular LLMs, such as ChatGPT \cite{achiam2023gpt}, Llama \cite{touvron2023llama}, ChatGLM \cite{glm2024chatglm}, and Qwen \cite{yang2024qwen2}, have been widely applied and are continuously being updated and iterated upon. However, current LLMs would exhibit issues such as hallucinations, inaccuracies, and challenges in disambiguating polysemous terms \cite{zhang2023siren, yang2024enhancing}. Therefore, it is crucial to develop a method to guide LLMs in generating more accurate and informative textual information. 

Recently, numerous works related to prompt engineering have emerged to fully harness the potential of LLMs, including approaches such as Chain of Thought \cite{wei2022chain}, In-Context Learning \cite{dong2022survey}, and Retrieval-Augmented Generation \cite{lewis2020retrieval}, which also inspire the design of the LENIE method. Additionally, the complementary strengths of LLMs and KGs have become a prominent research focus \cite{pan2024unifying}. KGs are derived from the real world, meaning that their nodes are generally recognizable by LLMs, enabling accurate description generation through prompt engineering to enhance semantic information in KGs.

In this work, we leverage LLMs is to generate more accurate and enriched descriptions for nodes in KGs, addressing the limitations posed by semantic deficiencies on NIE performance. Unlike existing approaches, we propose a clustering-based triplet sampling strategy to extract more informative, node-specific semantic information from KGs. This approach not only enables LLMs to understand the contextual information of nodes within KGs but also ensures that LLMs generate descriptions that are both accurate and information-rich. These augmented descriptions provide a more comprehensive and informative contents of nodes, thereby enhancing the performance of NIE on KGs.

\section{Methodology}\label{methodology}

\begin{figure*}[htbp]
    \centering
    \includegraphics[scale=0.34]{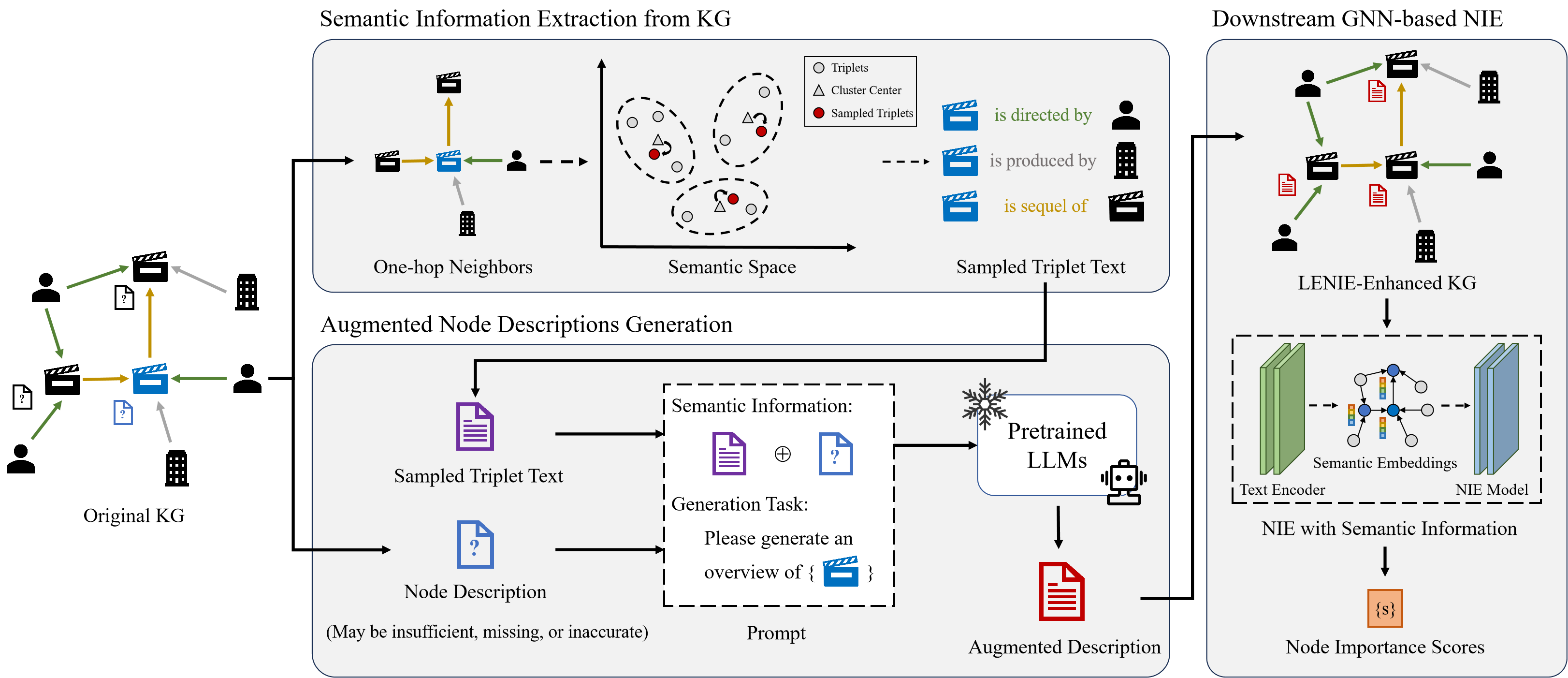}
    \caption{The overview of the proposed framework. LENIE extracts diverse semantic information from the given KG, generates augmented descriptions using LLMs, and encodes them into semantic embeddings to enhance downstream NIE performance.}
    \label{Fig:Overview}
\end{figure*}

\subsection{Problem Formulation}

\textbf{Knowledge Graphs or KGs}: A knowledge graph $\mathcal{G}=\mathcal{(V, R, T)}$, represents relationships between real-world entity nodes, where $\mathcal{V}$, $\mathcal{R}$ and $\mathcal{T}$ denote entities, relationships and triplets, respectively. Specifically, $(h,r,t)\in\mathcal{T}$ denotes a set of triplets, where head entity $h$ and tail entity $t$ both come from entity set $\mathcal{V}$, and relation $r$ comes from relation set $\mathcal{R}$. In a KG, which is a type of heterogeneous graph, two nodes can be connected by different types of edges, indicating that there can be multiple relationships between two entities.

\textbf{Semantic Information in KGs}: The defining characteristic of KGs is their rich semantic information, where each node can have its corresponding text description, and the triplets also reflect semantic information to some extent.

\textbf{Node Importance}: The importance of node $I_i \in \mathbb{R}^+$ is typically represented as a non-negative number, reflecting the significance of the entity within the KG. The value of $I_i$ is derived from real-world scores, such as movie ratings or the popularity of singers. Following previous works \cite{park2019estimating, huang2021representation, zhang2024label}, we utilize the log transformation of real-world scores as $I_i$.

\textbf{Node Importance Estimation or NIE}: Given a Knowledge graph $\mathcal{G}=\mathcal{(V, R, T)}$, a set of interested nodes $\mathcal{V}_s \subset \mathcal{V}$, and a set of partially known importance scores $\left \{ s \right \}$  for $\mathcal{V}_s$, NIE aims to learn a function $ f:\mathcal{V}_s \mapsto \mathbb{R}^+$ that predicts the node importance score for each node in $\mathcal{V}_s$. To ensure the comparability of the importance scores for nodes in $\mathcal{V}_s$, the nodes are typically of the same or similar type.

\subsection{Overview}
Figure \ref{Fig:Overview} illustrates the overview of the proposed framework, consisting of three key steps: Semantic Information Extraction from KG, Augmented Node Descriptions Generation, and NIE using downstream GNN-based models.

Specifically, we first develop a clustering-based triplet sampling strategy to effectively extract diverse semantic information for each interested node from KGs and integrate this information into triplet text. Next, we design node-specific adaptive prompts by combining the sampled triplet text with the node descriptions (if available) to define the nodes, enabling LLMs to generate accurate and informative augmented node descriptions, thus fully utilizing LLMs' understanding of real-world entites. Finally, the generated augmented node descriptions are encoded into semantically rich embeddings to further improve the performance of downstream NIE models. The details are elaborated in the following sections.

\subsection{Semantic Information Extraction from KG}\label{Semantic Information Extraction in KG}
Each entity node in a KG is connected to relevant entity nodes through triplets, which serve as the fundamental units representing the structure of the KG and inherently contain rich semantic information. To effectively extract this semantic information, we first perform subgraph extraction for each interested node $v_i \in \mathcal{V}_s$, i.e., we collect all triplets that involve $v_i$, as described below:
\begin{equation}
    \mathcal{T}_{v_i} = \left\{ (h, r, t) \mid \{h, t\} = \{v_i, n\}, \, v_i \in \mathcal{V}_s, \, n \in N_1(v_i), \, r \in \mathcal{R} \right\}
    \label{eq1}
\end{equation}
where $N_1(v_i)$ represents the set of all One-hop neighboring nodes of $v_i$, and $\{h, t\} = \{v_i, n\}$ indicates that if either $h$ or $t$ is $v_i$, the other node in the triplet is the neighboring node $n \in N_1(v_i)$, with $r \in \mathcal{R}$ denoting the relationship between $h$ and $t$. Note that the collected triplets must include the name texts of all $h$, $t$, and $r$ in $\mathcal{T}_{v_i}$. Subsequently, to better reflect the semantic information, we convert each triplet in $\mathcal{T}_{v_i}$ into its corresponding sentence text format, as follows:
\begin{equation}
    T_{v_i} = \left\{ f_\text{sentence}(h, r, t) \mid (h, r, t) \in \mathcal{T}_{v_i} \right\}
    \label{eq2}
\end{equation}
where $f_\text{sentence}$ is a function that converts a triplet $(h, r, t)$ into a sentence text, such as '$h$'s $r$ is $t$.' The above strategy extracts the semantic information of node $v_i$ from the KG in textual form. However, in real-world KGs, some popular nodes with many neighbors have numerous 
associated triplets, leading to excessively long and redundant extracted text, which increases processing difficulty for language models. Therefore, sampling the triplets connected to node $v_i$ is crucial. The triplet sampling strategy can either utilize random-based triplet sampling strategy or the proposed clustering-based triplet sampling strategy, with the differences explained as follows.

\textit{1) Random-based Triplet Sampling Strategy:} To ensure sampling fairness, random-based sampling with equal probability remains one of the most widely used algorithms. This strategy performs random-based triplet sampling without replacement from a large set of triplets, ensuring both fairness and uniqueness of the selected samples, as illustrated below:
\begin{equation}
    T_{v_i}' = \text{RandomSample}(T_{v_i}, \min(k, |T_{v_i}|))
    \label{eq3}
\end{equation}
where $T_{v_i}'$ represents the set of triplet sentences after random-based triplet sampling, $k$ is the desired number of sampled triplets, and $\min(k, |T_{v_i}|)$ ensures that the number of selected samples does not exceed the total number of triplet sentences $|T_{v_i}|$ in $T_{v_i}$. However, the random-based triplet sampling strategy is influenced by sample distribution. For instance, if $v_i$ shares the same relationship type with many neighboring nodes, triplet sentences associated with that relationship are more likely to be selected. In fact, we aim to extract semantic information about $v_i$ from a diverse range of relationship types or neighboring node types. Therefore, we recommend an alternative sampling strategy based on semantic clustering.

\textit{2) Clustering-based Triplet Sampling Strategy:} To extract semantic information from the KG that covers as many relationship and node types as possible, we propose clustering the triplet sentences in the semantic space for sampling. Specifically, all triplet sentences associated with $v_i$ are mapped into the semantic space using a text encoder, as follows:
\begin{equation}
    E_{v_i} = \left\{ f_\text{text\_encoder}(s) \mid s \in T_{v_i} \right\}
    \label{eq4}
\end{equation}
where $E_{v_i}$ represents the set of embeddings with semantic information obtained by encoding each triplet sentence $s$ in $T_{v_i}$. Next, the clustering algorithm is applied to find the cluster centers of the embeddings, as illustrated below:
\begin{equation}
    \{ c_1, c_2, \ldots, c_k \} = f_\text{cluster}(E_{v_i}, k)
    \label{eq5}
\end{equation}
where $c_k$ represents the $k$-th cluster center, and $k$ is the total number of clusters. Note that $k$ also cannot exceed the total number of triplet sentences $|T_{v_i}|$ in $T_{v_i}$. The clustering algorithm divides all triplet sentences related to $v_i$ into $k$ clusters, with the cluster centers representing the most distinct points in the semantic space. Therefore, selecting the triplet sentences closest to each cluster center in the semantic space is most likely to capture diverse relationship types and neighboring node types, i.e.,
\begin{equation}
    T_{v_i}' = \left\{ s_{m_j} \mid m_j = \arg\min_{m} d(c_j, e_m), \, j = 1, 2, \ldots, k \right\}
    \label{eq6}
\end{equation}
where $T_{v_i}'$ represents the set of triplet sentences after clustering sampling, $d(c_j, e_m)$ is the distance between cluster center $c_j$ and the embedding $e_m \in E_{v_i}$, and $s_{m_j} \in T_{v_i}$ is the triplet sentence corresponding to $e_{m}$ that is closest to $c_j$. 

Through the above strategy, we achieve semantic information extraction for each interested node in the KG by converting it into triplet sentence form. However, this semantic information is dependent on the scale and quality of the KG. To gain a more comprehensive understanding of these real-world nodes, we leverage LLMs, which possess vast knowledge bases, to generate text-based outputs that enhance the semantic information of these nodes.

\subsection{Augmented Node Descriptions Generation}

Leveraging the semantic information extracted from the KG, LLMs can mitigate the negative effects of hallucination and synonym ambiguity, enabling the generation of more accurate and informative description texts for real-world nodes, thereby enhancing the KG's overall semantic information to support the NIE task. To fully leverage the LLMs' understanding of $v_i$, we need to construct adaptive prompts tailored to different $v_i$ by incorporating its semantic information. After applying the strategy described in Section \ref{Semantic Information Extraction in KG}, we concatenate all the sampled triplet sentences of $v_i$ into a single text, as follows:
\begin{equation}
    T_{v_i}^{\text{Triplets}} = \bigcup_{s \in T_{v_i}'}s
    \label{eq7}
\end{equation}
where $T_{v_i}^{\text{Triplets}}$ represents the sampled triplet text for node $v_i$, which includes the triplet sentences obtained through sampling. Note that the length of $T_{v_i}^{\text{Triplets}}$ must remain below the input limit of the LLM, allowing room for additional content. Next, we combine all available semantic information of $v_i$ from the KG, i.e., the sampled triplet text and the original node description text (if available), to construct the adaptive prompt, as detailed below:
\begin{equation}
    \text{prompt}_{v_i} = T_{v_i}^{\text{Triplets}} \cup T_{v_i}^{\text{Descriptions}} \cup T^{\text{Generation\_Task}}
    \label{eq8}
\end{equation}
where $\text{prompt}_{v_i}$ is composed of $v_i$'s sampled triplet text $T_{v_i}^{\text{Triplets}}$, description text $T_{v_i}^{\text{Descriptions}}$, and $T^{\text{Generation\_Task}}$ which provides the instruction for the LLMs to perform augmented description text generation. Each interested node $v_i$'s augmented description text $D_{v_i}^{\text{Generated}}$ is generated by the LLM, guided by its corresponding prompt, as follows:
\begin{equation}
    D_{v_i}^{\text{Generated}} = \text{LLM}(\text{prompt}_{v_i}) .
    \label{eq9}
\end{equation}

The augmented description text generated by the above method not only incorporates the semantic information of the node from the KG and its own textual description (if available) but also integrates external knowledge provided by the LLM. Additionally, this approach addresses semantic deficiencies in KGs, enhancing the node's semantic representation and improving NIE performance in downstream GNN-based models.

\subsection{Downstream GNN-based NIE}

To assess node importance on KGs with diverse node and edge types, we utilize a GNN-based model that integrates both structural and augmented semantic information, aiming to achieve superior performance in the NIE task. Augmented descriptions $D_i^{\text{Generated}}$ for each $v_i \in V_s$ are generated using the proposed method, then encoded into vectors via a text encoder as their initial embeddings $h_i^{(0)}$, as shown below:
\begin{equation}
    h_i^{(0)} = f_{\text{text\_encoder}}(D_i^{\text{Generated}}), \quad \forall v_i \in V_s .
    \label{eq10}
\end{equation}
These $h_i^{(0)}$ capture rich semantic information. Subsequently, the downstream GNN-based NIE model aggregates and updates these information across the graph, as described below:
\begin{equation}
    h_{i}^{(t+1)} = \text{Update}\left( h_i^{(t)}, \bigoplus_{j \in \mathcal{N}(i)} \psi(h_i^{(t)}, h_j^{(t)}, h_{ij}^{(t)}) \right) 
    \label{eq11}
\end{equation}
Where $h_i^{(t)}$ and $h_j^{(t)}$ represent the embeddings of node $v_i$ and its neighboring node $j \in \mathcal{N}(i)$ at time step $t$, and $h_{ij}^{(t)}$ is the embedding of the edge between them. The GNN-based model first aggregates information from $v_i$, its neighbors, and the corresponding edges by applying differentiable functions $\psi$ (e.g., artificial neural networks)
and an information aggregation operator $\bigoplus$ (e.g., sum, mean, or max), to compute the message around $v_i$ in the graph. Subsequently, the model updates $v_i$'s embedding by combining its $h_i^{(t)}$ with the message, resulting in the updated embedding $h_i^{(t+1)}$ at time step $t+1$. This approach enables the GNN-based model to facilitate feature interactions across various node and edge types, enhancing performance in the NIE task.

Currently, the State-Of-The-Art (SOTA) performance in NIE tasks on KGs is achieved by RGTN \cite{huang2021representation} and its improved version, LICAP \cite{zhang2024label}, both of which are GNN-based models. Our proposed semantic enhancement method LENIE can further improve their performance. Specifically, RGTN aggregates and updates the LENIE-enhanced embeddings $h^{\text{LENIE}}$ on the graph as $h^{(t)}=f_{\text{GNN-RGTN}}(h^\text{{LENIE}})$, while using projection matrices to compute node importance scores $s_i = f_{\text{scores-RGTN}}(h^{(t)})$ for each $v_i$. Building on this, LICAP introduces contrastive learning for pretraining embeddings, aiming to make nodes with similar importance scores have closer embeddings, expressed as $h^{\text{LICAP}}=f_{\text{pretraining-LICAP}}(h^{\text{LENIE}})$ thereby further enhancing the performance of downstream RGTN model. LENIE’s focus on semantic information enhancement not only improves the performance of current SOTA models on the NIE task but also benefits a wide range of downstream NIE models. This suggests that LENIE could continue to enhance NIE models in the future, enabling more accurate node importance prediction on KGs.

\section{Experimental Settings}\label{experimental settings}

\subsection{Dataset}

\begin{table}[htbp]
  \centering
  \caption{\centering{The statistics of real-world knowledge graphs.}}
    \scalebox{0.93}{
    \begin{tabular}{lrrrr}
    \toprule
    Datasets & Nodes & Edges & Relationships & Nodes with Importance \\
    \midrule
    FB15K & 14951 & 592213 & 1345  & 14105 (94.3\%) \\
    TMDB5K & 114805 & 761648 & 34    & 4803 (4.2\%) \\
    MUSIC10K & 22985 & 65290 & 8     & 4412 (19.2\%) \\
    \bottomrule
    \end{tabular}
    }
  \label{tab_statistics1}
\end{table}
To assess the effectiveness of the proposed method, we perform experiments on three real-world knowledge graphs. The statistics of them are shown in Table \ref{tab_statistics1}.
Each dataset includes node importance scores, triplets, and textual information. The logarithm of node importance scores are used as ground truth labels for the experiments.
\begin{itemize}
\item \textbf{FB15K} is a subset of Freebase\footnote{\url{http://www.freebase.be/}}. It comprises 14,951 entities and 1,345 distinct relationships, reflecting the world's knowledge through triplets. The pageviews of Wikipedia pages are used as the nodes' importance.
\item \textbf{TMDB5K} is a movie knowledge graph generated from TMDB 5000 dataset\footnote{\url{https://www.kaggle.com/tmdb/tmdb-movie-metadata}}. It includes around 5,000 movie entities, along with entities such as actors, directors, and countries.  The popularity scores of movies function as the movie nodes' importance. 
\item \textbf{MUSIC10K} is a music knowledge graph constructed from the 10k Song Dataset\footnote{\url{https://www.kaggle.com/datasets/mexwell/10k-song-dataset}}, with additional information sourced from the Million Song Dataset\footnote{\url{https://millionsongdataset.com/}}. It contains about 22985 entities such as artists, songs, terms of artists and releases. This dataset lacks node description text, so entity names serve as it. The familiarity of artists is defined as the artist nodes' importance.
\end{itemize}

We thank Chen et al. \cite{chen2024deep} for generously providing the original MUSIC10K dataset. MUSIC10K is further processed into the similar format as used in FB15K and TMDB5K, and we make the well-processed MUSIC10K freely available at \url{https://github.com/XinyuLin-FZ/LENIE}. Regarding FB15K and TMDB5K datasets, we directly obtain them from \url{https://github.com/GRAPH-0/RGTN-NIE}.

\subsection{Baseline Methods}
To comprehensively evaluate the advantages that integrating LENIE brings to NIE models and to further clarify the importance of enhanced semantic information for NIE tasks, we follow \cite{zhang2024label} by comparing three distinct types of NIE methods. These three types of NIE methods, covering a total of 10 NIE models, are used as baselines for comparison.
\begin{enumerate}[topsep=0pt, parsep=0pt, itemsep=0pt, partopsep=0pt,label=\arabic*)]
    \item \textit{Unsupervised Methods:} PR \cite{page1999pagerank} and PPR \cite{haveliwala2002topic} are two well-known methods that assess node importance based on graph topology, without requiring labels or node features for training.
    \item \textit{Non-GNN Supervised Methods:} Treating NIE as a regression task, we can utilize labels and node features to train two classical regression models—Linear Regression (LR) and MultiLayer Perception (MLP)—to predict node importance scores.
    \item \textit{GNN-based Supervised Methods:} These methods leverage labels, structural, and semantic information from KGs to train the learning model. 
    
    We employ general-purpose GNN-based models, such as GCN \cite{kipf2016semi}, GraphSAGE \cite{hamilton2017inductive}, and RGCN \cite{schlichtkrull2018modeling}, tailored for the NIE task, along with SOTA NIE-specific GNN-based models like GENI \cite{park2019estimating}, RGTN \cite{huang2021representation}, and LICAP \cite{zhang2024label}, all serving as benchmarks for comparison.

\end{enumerate}

\subsection{Evaluation Metrics}
To comprehensively evaluate the performance of the NIE models, we employ both regression metrics and ranking metrics, totaling 5 metrics, consistent with \cite{zhang2024label}. The specific definitions of these metrics are as follows:
\begin{enumerate}[topsep=0pt, parsep=0pt, itemsep=0pt, partopsep=0pt,label=\arabic*)]
    \item \textit{Regression metrics:} The NIE model's predicted scores align with nodes' ground-truth labels, both reflecting node importance in KGs as numerical values. Therefore, we employ regression metrics, specifically Root Mean Square Error (RMSE) and Median Absolute Error (MedianAE), to evaluate the discrepancy between predicted and actual scores. Lower values indicate better performance.
    
    \item \textit{Ranking metrics:} To evaluate the performance from another perspective, comparing the node importance rankings derived from predicted scores ${\left\{ s_i^* \right\}}$ and ground-truth scores ${\left\{ s_i \right\}}$ can also effectively assesses the alignment between the NIE model’s predictions and actual results. Therefore, we employ three ranking metrics: Normalized Discounted Cumulative Gain (NDCG), Spearman's rank correlation coefficient (SPEARMAN), and Overlap (OVER). Higher values indicate better performance.
    
\end{enumerate}

\subsection{Implementation Details}

LENIE aims to enhance semantic information in KGs to generate semantic embeddings for nodes, thereby improving downstream NIE model performance. In the subsequent experiments, unless otherwise specified, LENIE adopts the recommended clustering-based triplet sampling strategy for semantic extraction, selecting 10 triplets for FB15K, 5 for TMDB5K, and 3 for MUSIC10K according to their average node degrees, with Llama3.1\footnote{\url{https://github.com/meta-llama/llama-models/tree/main}} as the default LLM to generate augmented descriptions of nodes. Throughout the process, the all-mpnet-base-v2\footnote{\url{https://huggingface.co/sentence-transformers/all-mpnet-base-v2}} (which maps text to a 768-dimensional dense vector space) serves as the text encoder. In all experiments, node original descriptions are encoded and fed into downstream NIE models, with their performance serving as a baseline for comparison. This setup demonstrates the effectiveness of LENIE’s semantic enhancement of KGs in boosting the performance of downstream NIE models.

All techniques and models employed in the experiments are implemented in Python. Most experimental settings align with \cite{zhang2024label}, with specific configurations as follows: PR and PPR are from the NetworkX package\footnote{\url{https://networkx.org/}}, LR, MLP, and K-means from scikit-learn package\footnote{\url{https://scikit-learn.org/stable/}}, and GCN, GraphSAGE, and RGCN from the DGL package\footnote{\url{https://www.dgl.ai/}}, each with a hidden layer of 64 dimensions. The LICAP model is sourced from its GitHub repository\footnote{\url{https://github.com/zhangtia16/LICAP}}. Note that both RGTN and LICAP achieve state-of-the-art performance, with LICAP serving as an improved version of RGTN by enhancing node embeddings through pretraining. To assess the impact of text information augmentation, RGTN is used as the downstream NIE model for LENIE in the main experiments, providing a comparison against other benchmark methods. For all experiments, we conducted a learning rate search across the range [0.1, 0.5, 0.01, 0.05, 0.001, 0.005, 0.0001, 0.0005]. Other hyperparameters were set to each model’s default optimal values, and all models were evaluated using 5-fold cross-validation. The learning rate with the lowest RMSE was chosen as optimal, with corresponding performance results recorded as final outcomes. All experiments were conducted on NVIDIA 3090 GPUs.

\section{Experiments}\label{experiments}
In this section, we examine the effectiveness of LENIE and its components by addressing the following questions.

\begin{enumerate}[topsep=0pt, parsep=0pt, itemsep=0pt, partopsep=0pt,label=\arabic*)]
    \item Can the proposed LENIE further boost the performance of existing NIE model and achieve the new state-of-the-art performance? (Section \ref{RQ1})
    \item How do different semantic augmentation approaches affect the NIE performance? (Section \ref{RQ2})
    \item Why is the proposed clustering-based triplet sampling superior to random-based sampling? (Section \ref{RQ3})
    \item Do the augmented descriptions generated by different LLMs improve the NIE performance? (Section \ref{RQ4})
    \item Can the proposed LENIE framework also improve the performance of various NIE models? (Section \ref{RQ5})
    \item How are the semantic deficiencies in KGs, such as insufficient descrpitons, missing descriptions, and inaccurate information, addressed by LENIE?
    (Section \ref{RQ6})
    
\end{enumerate}

\subsection{NIE on Real-World KGs} \label{RQ1}

The main experimental results of our proposed LENIE method compared to other NIE methods across three real-world KGs are presented in Table \ref{main exp}. Conclusions drawn from these results can be explained from the following perspectives.

LENIE achieves state-of-the-art performance across nearly all metrics in NIE tasks on three real-world KGs. On the FB15K dataset, LENIE’s OVER@100 is slightly lower than RGTN by 0.4\%, but it shows an average improvement of approximately 0.7\% across other metrics, with a maximum increase of 1\% in RMSE. On the TMDB5K dataset, LENIE outperforms RGTN across all metrics with an average improvement of around 3.7\%, with the largest gain of 6\% in OVER@100. Similarly, on the MUSIC10k dataset, LENIE excels across all metrics, showing an average improvement of 6.4\% over RGTN and a substantial 19.6\% improvement in SPEARMAN.

Effectively leveraging additional node information in KGs enhances the performance of NIE models. Unsupervised methods like PR and PPR rely solely on graph topology for unsupervised training and generally perform worse on NIE metrics across the three real-world KGs compared to other methods. Non-GNN supervised methods like LR and MLP utilize labels and node features, achieving better overall performance than unsupervised methods; however, they cannot fully exploit the rich information within KGs, limiting further improvement. In contrast, GNN-based supervised methods capture rich structural and semantic information through aggregation and update mechanisms, yielding better overall performance and making it easier to achieve state-of-the-art results.

Enhancing the semantic information of nodes in KGs is essential for improving NIE model performance. While FB15K and TMDB5K datasets include node original descriptions, MUSIC10K dataset contains only node names, providing limited semantic context. LENIE bridges this gap by integrating semantic information from KGs with LLMs to generate augmented descriptions, significantly enriching node semantics. This enhancement explains LENIE’s especially strong impact on downstream RGTN models in MUSIC10K compared to other KGs. These semantic improvements also benefit other NIE models, as discussed further in Section \ref{RQ5}.    

\begin{table*}[htbp]
  \centering
  \caption{\centering{The experimental results of various NIE models across three real-world KGs using five metrics. Arrows indicate trends: down means better performance with lower values, up means better performance with higher values. Bold text highlights the best-performing model for each metric, with the second-best underlined.}}
    \begin{tabular}{llllll}
    \toprule
          & \multicolumn{5}{c}{FB15K} \\
\cmidrule{2-6}    Methods & RMSE $\downarrow$ & MedianAE $\downarrow$ & NDCG@100 $\uparrow$ & SPEARMAN $\uparrow$ & OVER@100 $\uparrow$ \\
    \midrule
    PR    & 9.6920±0.2006 & 9.6025±0.0468 & 0.8400±0.0103 & 0.3497±0.0188 & 0.1520±0.0172 \\
    PPR   & 9.6905±0.2041 & 9.5948±0.0474 & 0.8411±0.0112 & 0.3500±0.0192 & 0.1540±0.0185 \\
    \midrule
    LR    & 0.9051±0.0137 & 0.5515±0.0110 & 0.9407±0.0076 & 0.7837±0.0044 & 0.3880±0.0232 \\
    MLP   & 1.1125±0.0163 & 0.6926±0.0134 & 0.9246±0.0065 & 0.7075±0.0109 & 0.3320±0.0387 \\
    \midrule
    GCN   & 2.4122±0.3064 & 1.5846±0.3339 & 0.8778±0.0147 & 0.4660±0.0288 & 0.1800±0.0352 \\
    RGCN  & 4.0882±1.5456 & 2.5390±0.9453 & 0.8547±0.0189 & 0.2859±0.0264 & 0.1620±0.0608 \\
    GraphSAGE & 0.8645±0.0156 & 0.5136±0.0103 & 0.9491±0.0081 & 0.8076±0.0088 & 0.4380±0.0160 \\
    GENI  & 0.9456±0.0295 & 0.5487±0.0165 & 0.9174±0.0139 & 0.7710±0.0094 & 0.3260±0.0706 \\
    LICAP & 0.9373±0.0578 & 0.5596±0.0431 & 0.9409±0.0086 & 0.7839±0.0130 & 0.4040±0.0546 \\
    \midrule
    RGTN  & \underline{0.7926±0.0176} & \underline{0.4540±0.0080} & \underline{0.9587±0.0034} & \underline{0.8370±0.0101} & \textbf{0.5020±0.0194} \\
    LENIE(Ours) & \textbf{0.7826±0.0158} & \textbf{0.4468±0.0075} & \textbf{0.9629±0.0037} & \textbf{0.8427±0.0036} & \underline{0.4980±0.0331} \\
    \midrule
          & \multicolumn{5}{c}{TMDB5K} \\
\cmidrule{2-6}    Methods & RMSE $\downarrow$ & MedianAE $\downarrow$ & NDCG@100 $\uparrow$ & SPEARMAN $\uparrow$ & OVER@100 $\uparrow$ \\
    \midrule
    PR    & 2.1411±0.0114 & 2.0025±0.0371 & 0.8387±0.0102 & 0.6247±0.0130 & 0.4100±0.0616 \\
    PPR   & 1.8538±0.0101 & 1.6869±0.0241 & 0.8495±0.0083 & 0.6856±0.0098 & 0.4200±0.0506 \\
    \midrule
    LR    & 1.1427±0.0217 & 0.7688±0.0341 & 0.7507±0.0337 & 0.3569±0.0321 & 0.3300±0.0303 \\
    MLP   & 1.2940±0.0180 & 0.9049±0.0268 & 0.7192±0.0270 & 0.2751±0.0100 & 0.2680±0.0397 \\
    \midrule
    GCN   & 0.9656±0.0486 & 0.6323±0.0288 & 0.8620±0.0233 & 0.6588±0.0471 & 0.4480±0.0417 \\
    RGCN  & 0.8703±0.0207 & 0.5926±0.0151 & 0.8656±0.0171 & 0.7282±0.0240 & 0.4640±0.0314 \\
    GraphSAGE & 0.9897±0.0159 & 0.6629±0.0278 & 0.8297±0.0322 & 0.5467±0.0318 & 0.4160±0.0432 \\
    GENI  & 0.7741±0.0291 & 0.5039±0.0326 & 0.8522±0.0132 & 0.7345±0.0219 & 0.4240±0.0492 \\
    LICAP & 0.7861±0.0161 & 0.5116±0.0358 & 0.8851±0.0195 & 0.7306±0.0178 & 0.5180±0.0172 \\
    \midrule
    RGTN  & \underline{0.7607±0.0133} & \underline{0.4935±0.0173} & \underline{0.8996±0.0191} & \underline{0.7480±0.0164} & \underline{0.5340±0.0242} \\
    LENIE(Ours) & \textbf{0.7188±0.0312} & \textbf{0.4635±0.0165} & \textbf{0.9133±0.0118} & \textbf{0.7851±0.0167} & \textbf{0.5940±0.0242} \\
    \midrule
          & \multicolumn{5}{c}{MUSIC10K} \\
\cmidrule{2-6}    Methods & RMSE $\downarrow$ & MedianAE $\downarrow$  & NDCG@100 $\uparrow$ & SPEARMAN $\uparrow$ & OVER@100 $\uparrow$ \\
    \midrule
    PR    & 0.4897±0.0116 & 0.1955±0.0077 & 0.7992±0.0140 & 0.1772±0.0189 & 0.2360±0.0294 \\
    PPR   & 0.5562±0.0039 & 0.5037±0.0030 & 0.7981±0.0109 & 0.1890±0.0228 & 0.2380±0.0299 \\
    \midrule
    LR    & 0.1095±0.0047 & 0.0652±0.0014 & 0.8702±0.0171 & 0.3566±0.0199 & 0.4420±0.0445 \\
    MLP   & 0.4606±0.0342 & 0.3045±0.0196 & 0.7229±0.0244 & 0.0349±0.0523 & 0.1500±0.0237 \\
    \midrule
    GCN   & 0.1049±0.0030 & 0.0589±0.0016 & 0.8854±0.0199 & 0.4440±0.0326 & 0.4680±0.0387 \\
    RGCN  & 0.0972±0.0037 & 0.0565±0.0016 & 0.8974±0.0087 & \underline{0.5377±0.0097} & 0.4880±0.0232 \\
    GraphSAGE & 0.0983±0.0039 & 0.0579±0.0020 & 0.9076±0.0111 & 0.4591±0.0195 & 0.5240±0.0338 \\
    GENI  & 0.0985±0.0045 & 0.0597±0.0021 & 0.8872±0.0065 & 0.4928±0.0262 & 0.4780±0.0194 \\
    LICAP & 0.0986±0.0045 & 0.0598±0.0029 & \underline{0.9130±0.0133} & 0.4717±0.0331 & \underline{0.5420±0.0397} \\
    \midrule
    RGTN  & \underline{0.0970±0.0042} & \underline{0.0564±0.0018} & 0.9126±0.0114 & 0.4590±0.0237 & 0.5360±0.0326 \\
    LENIE(Ours) & \textbf{0.0856±0.0035} & \textbf{0.0478±0.0018} & \textbf{0.9395±0.0032} & \textbf{0.6553±0.0164} & \textbf{0.6120±0.0279} \\
    \bottomrule
    \end{tabular}
  \label{main exp}
\end{table*}

\subsection{Ablation Study: Semantic Augmentation} \label{RQ2}
The semantic augmentation process in LENIE involves semantic extraction from KGs via the clustering-based triplet sampling strategy, followed by prompting LLMs to generate augmented descriptions for nodes in the KGs. To investigate the effectiveness of these designs, we conducted an ablation study, with results presented in Table \ref{exp_method}. The relevant definitions are as follows:
\begin{itemize}
    \item \textbf{Vanilla}: The original RGTN method without LENIE.
    \item \textbf{LENIE (concate)}: LENIE enhances node semantics by concatenating node original description with triplet text extracted from KGs, without using LLMs.
    \item \textbf{LENIE (random)}: LENIE enhances node semantics by integrating triplet text extracted via random-based triplet sampling strategy with the original node descriptions to prompt LLMs in generating augmented descriptions.
    \item \textbf{LENIE (cluster)}: Building on the previous approach, LENIE applies the clustering-based triplet sampling strategy to extract semantic information from KGs.
\end{itemize}

As shown in Table \ref{exp_method}, LENIE (concate), which incorporates semantic extraction from KGs, outperforms the Vanilla method (which uses only the original node descriptions) on both TMDB5K and MUSIC10K datasets across most metrics. Furthermore, the LENIE (random) and LENIE (cluster) methods, which additionally leverage LLMs for augmented text generation, achieve overall superior performance across all KGs compared to the Vanilla method. These results validate the effectiveness of both semantic extraction from KGs and LLM-based text generation.

Moreover, comparing the performance of LENIE (random) and LENIE (cluster) reveals that clustering-based triplet sampling strategy for semantic extraction, followed by LLMs augmented text generation, is more effective than random-based strategy for enhancing NIE model performance. The ablation experiment results above validate the effectiveness of LENIE and its semantic augmentation method.

\begin{table*}[htbp]
  \centering
  \caption{\centering{The performance of LINIE with different semantic augmentation approaches.}}
    \begin{tabular}{llllll}
    \toprule
          & \multicolumn{5}{c}{FB15K} \\
\cmidrule{2-6}    Methods & \multicolumn{1}{l}{RMSE $\downarrow$} & \multicolumn{1}{l}{MedianAE $\downarrow$} & \multicolumn{1}{l}{NDCG@100 $\uparrow$} & \multicolumn{1}{l}{SPEARMAN $\uparrow$} & \multicolumn{1}{l}{OVER@100 $\uparrow$} \\
    \midrule
    Vanilla & 0.7926±0.0176 & \underline{0.4540±0.0080} & 0.9587±0.0034 & 0.8370±0.0101 & \textbf{0.5020±0.0194} \\
     \textbf{+LENIE} (concat) & 0.8213±0.0105 & 0.4889±0.0122 & 0.9534±0.0025 & 0.8248±0.0042 & 0.4440±0.0377 \\
     \textbf{+LENIE} (random) & \underline{0.7887±0.0157} & 0.4611±0.0139 & \underline{0.9607±0.0025} & \underline{0.8407±0.0082} & 0.4920±0.0487 \\
     \textbf{+LENIE} (cluster) & \textbf{0.7826±0.0158} & \textbf{0.4468±0.0075} & \textbf{0.9629±0.0037} & \textbf{0.8427±0.0036} & \underline{0.4980±0.0331} \\
    \midrule
          & \multicolumn{5}{c}{TMDB5K} \\
\cmidrule{2-6}    Methods & \multicolumn{1}{l}{RMSE $\downarrow$} & \multicolumn{1}{l}{MedianAE $\downarrow$} & \multicolumn{1}{l}{NDCG@100 $\uparrow$} & \multicolumn{1}{l}{SPEARMAN $\uparrow$} & \multicolumn{1}{l}{OVER@100 $\uparrow$} \\
    \midrule
    Vanilla & 0.7607±0.0133 & 0.4935±0.0173 & 0.8996±0.0191 & 0.7480±0.0164 & 0.5340±0.0242 \\
     \textbf{+LENIE} (concat) & 0.7496±0.0319 & 0.4927±0.0293 & \textbf{0.9160±0.0093} & 0.7586±0.0354 & \textbf{0.5960±0.0162} \\
     \textbf{+LENIE} (random) & \underline{0.7205±0.0231} & \underline{0.4689±0.0256} & 0.9123±0.0092 & \underline{0.7845±0.0148} & 0.5940±0.0393 \\
     \textbf{+LENIE} (cluster) & \textbf{0.7188±0.0312} & \textbf{0.4635±0.0165} & \underline{0.9133±0.0118} & \textbf{0.7851±0.0167} & \underline{0.5940±0.0242} \\
    \midrule
          & \multicolumn{5}{c}{MUSIC10K} \\
\cmidrule{2-6}    Methods & \multicolumn{1}{l}{RMSE $\downarrow$} & \multicolumn{1}{l}{MedianAE $\downarrow$} & \multicolumn{1}{l}{NDCG@100 $\uparrow$} & \multicolumn{1}{l}{SPEARMAN $\uparrow$} & \multicolumn{1}{l}{OVER@100 $\uparrow$} \\
    \midrule
    Vanilla & 0.0970±0.0042 & 0.0564±0.0018 & 0.9126±0.0114 & 0.4590±0.0237 & 0.5360±0.0326 \\
     \textbf{+LENIE} (concat) & 0.0949±0.0040 & 0.0558±0.0008 & 0.9099±0.0107 & 0.5071±0.0088 & 0.5280±0.0412 \\
     \textbf{+LENIE} (random) & \textbf{0.0855±0.0042} & \underline{0.0486±0.0013} & \underline{0.9385±0.0104} & \underline{0.6488±0.0191} & \underline{0.5920±0.0504} \\
     \textbf{+LENIE} (cluster) & \underline{0.0856±0.0035} & \textbf{0.0478±0.0018} & \textbf{0.9395±0.0032} & \textbf{0.6553±0.0164} & \textbf{0.6120±0.0279} \\
    \bottomrule
    \end{tabular}
  \label{exp_method}
\end{table*}

\subsection{Comparison of Triplet Texts Extracted by Two Strategies} \label{RQ3}

\begin{figure*}
    \centering
    \includegraphics[scale=0.46]{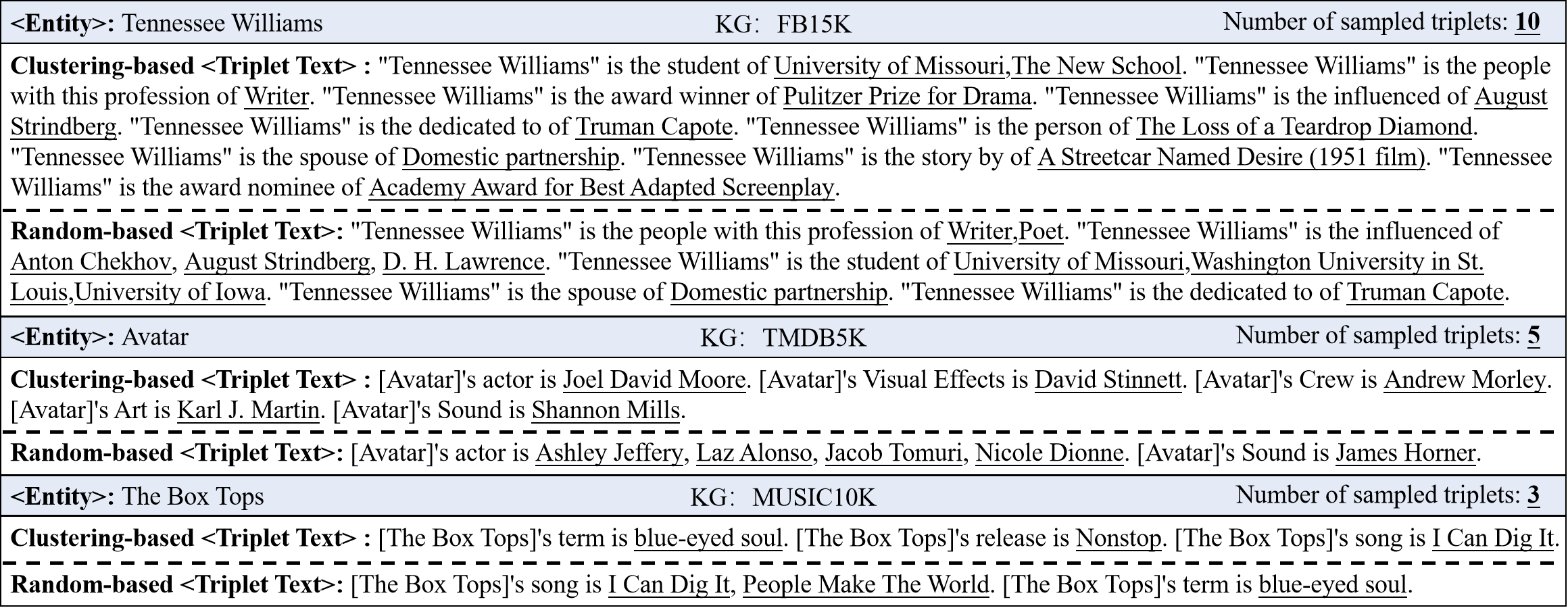}
    \caption{\centering{Comparison of triplet texts extracted by the clustering-based and random-based strategies. The clustering-based strategy extracts more comprehensive semantic information than the random-based strategy.}}
    \label{Fig:cr}
\end{figure*}

To further investigate why the clustering-based triplet sampling strategy in LENIE outperforms the random-based triplet sampling strategy for the NIE task, we conducted a textual comparison of the semantic information extracted for the same node using both strategies across three KGs.

As shown in Figure \ref{Fig:cr}, under the same sampling count, clustering-based triplet sampling strategy tends to select triplet sentences with the most diverse semantics. The resulting triplet text better guides LLMs in generating more comprehensive descriptions for nodes (details are provided in Section \ref{RQ6}). In contrast, the random-based triplet sampling strategy is more influenced by the distribution of nodes in KGs, often resulting in the collection of a large number of semantically similar triplets, thereby limiting semantic diversity. Consequently, clustering-based triplet sampling strategy enables the extraction of richer semantic information from KGs for nodes, facilitating a more holistic estimation of node importance.

\subsection{Impact of Different LLMs on LENIE} \label{RQ4}

\begin{table*}[htbp]
  \centering
  \caption{\centering{The performance of LENIE with different LLMs for semantic augmentation.}}
    \begin{tabular}{llllll}
    \toprule
          & \multicolumn{5}{c}{FB15k} \\
\cmidrule{2-6}    Methods & RMSE $\downarrow$ & MedianAE $\downarrow$ & NDCG@100 $\uparrow$ & SPEARMAN $\uparrow$ & OVER@100 $\uparrow$ \\
    \midrule
    Vanilla & \underline{0.7926±0.0176} & \underline{0.4540±0.0080} & 0.9587±0.0034 & 0.8370±0.0101 & \textbf{0.5020±0.0194} \\
     \textbf{+LENIE} (Llama3.1) & \textbf{0.7826±0.0158} & \textbf{0.4468±0.0075} & \textbf{0.9629±0.0037} & \textbf{0.8427±0.0036} & \underline{0.4980±0.0331} \\
     \textbf{+LENIE} (GLM4) & 0.7949±0.0232 & 0.4556±0.0124 & \underline{0.9595±0.0049} & 0.8375±0.0042 & 0.4900±0.0684 \\
     \textbf{+LENIE} (Qwen2) & 0.7934±0.0130 & 0.4545±0.0158 & 0.9574±0.0016 & \underline{0.8405±0.0102} & 0.4880±0.0483 \\
     \midrule
          & \multicolumn{5}{c}{TMDB5K} \\
\cmidrule{2-6}    Methods & RMSE $\downarrow$ & MedianAE $\downarrow$ & NDCG@100 $\uparrow$ & SPEARMAN $\uparrow$ & OVER@100 $\uparrow$ \\
    \midrule
    Vanilla & 0.7607±0.0133 & 0.4935±0.0173 & 0.8996±0.0191 & 0.7480±0.0164 & 0.5340±0.0242 \\
     \textbf{+LENIE} (Llama3.1) & \underline{0.7188±0.0312} & \underline{0.4635±0.0165} & 0.9133±0.0118 & \underline{0.7851±0.0167} & \underline{0.5940±0.0242} \\
     \textbf{+LENIE} (GLM4) & 0.7192±0.0110 & 0.4754±0.0195 & \underline{0.9161±0.0123} & 0.7841±0.0091 & 0.5880±0.0431 \\
     \textbf{+LENIE} (Qwen2) & \textbf{0.6918±0.0120} & \textbf{0.4398±0.0211} & \textbf{0.9178±0.0078} & \textbf{0.8020±0.0117} & \textbf{0.6060±0.0215} \\
    \midrule
          & \multicolumn{5}{c}{MUSIC10K} \\
\cmidrule{2-6}    Methods & RMSE $\downarrow$ & MedianAE $\downarrow$ & NDCG@100 $\uparrow$ & SPEARMAN $\uparrow$ & OVER@100 $\uparrow$ \\
    \midrule
    Vanilla & 0.0970±0.0042 & 0.0564±0.0018 & 0.9126±0.0114 & 0.4590±0.0237 & 0.5360±0.0326 \\
     \textbf{+LENIE} (Llama3.1) & \textbf{0.0856±0.0035} & \textbf{0.0478±0.0018} & \textbf{0.9395±0.0032} & \textbf{0.6553±0.0164} & \underline{0.6120±0.0279} \\
     \textbf{+LENIE} (GLM4) & 0.0887±0.0036 & 0.0504±0.0030 & 0.9323±0.0111 & 0.6060±0.0138 & 0.5740±0.0174 \\
     \textbf{+LENIE} (Qwen2) & \underline{0.0877±0.0039} & \underline{0.0496±0.0019} & \underline{0.9387±0.0061} & \underline{0.6162±0.0148} & \textbf{0.6140±0.0273} \\
    \bottomrule
    \end{tabular}
  \label{exp_llm}
\end{table*}

To assess whether LENIE based on different LLMs also improves NIE model performance, we compared LENIE's effectiveness using the latest versions of Llama, GLM, and Qwen (i.e., Llama3.1-8b-Instruct, GLM4-9b-Chat\footnote{\url{https://huggingface.co/THUDM/glm-4-9b-chat}}, and Qwen2-7b-Instruct\footnote{\url{https://huggingface.co/Qwen/Qwen2-7B-Instruct}}) for the NIE task. All other experimental settings align with Section \ref{RQ1}, except for the choice of LLMs.

As shown in Table \ref{exp_llm}, the experimental results indicate that LENIE with various LLMs enhances most performance metrics across KGs. Among these, Llama3.1 achieves the best overall performance across the three KGs, while Qwen2 performs best on the TMDB5K. GLM4, however, shows relatively lower overall performance. These performance differences among LLM-based LENIE methods may result from variations in training data and application domains across different LLMs. With the rapid advancement of LLMs, future models are expected to further enhance NIE model performance through the LENIE approach.

\subsection{Employing LENIE across Various NIE Methods} \label{RQ5}
To investigate whether LENIE's semantic augmentation of KGs improves performance across various NIE models, we evaluated seven models, including both Non-GNN based and GNN based approaches, on five NIE metrics across three KGs. Specifically, we initialized each NIE model's node embeddings with semantic embeddings derived from either the original node descriptions provided by the datasets or the LENIE-augmented descriptions, allowing us to assess the impact of enhanced KGs semantics on NIE tasks. All models were tuned for learning rates within the range [0.1, 0.5, 0.01, 0.05, 0.001, 0.005, 0.0001, 0.0005], with 5-fold cross validation used to select the best-performing configuration for presentation. LENIE applies our recommended clustering-based triplet sampling strategy to extract semantic information from KGs, with Llama3.1 used in the augmented description generation step. Experimental results are shown in Table \ref{exp_models}.

According to the experimental results shown in the table, LENIE’s semantic augmentation of KGs generally enhances most NIE metrics across all NIE models on various KGs. Several models even show substantial improvements in both regression and ranking metrics with LENIE’s support. Notably, for MUSIC10K, which lacks original node descriptions, LENIE achieves significant improvements across all NIE metrics for all models, with the exception of a slight decrease in GCN’s performance on the OVER@100 metric, though with lower performance variance. In summary, semantic information in KGs is crucial for NIE tasks, and LENIE effectively enriches this information—especially in KGs missing node descriptions—thereby boosting the performance of existing NIE models and supporting NIE tasks.

\begin{table*}[htbp]
  \centering
  \caption{\centering{The performance of LINIE for various downstream NIE models.}}
    \begin{tabular}{lllllll}
    \toprule
          &       & \multicolumn{5}{c}{FB15K} \\
\cmidrule{3-7}          & Methods & RMSE $\downarrow$ & MedianAE $\downarrow$ & NDCG@100 $\uparrow$ & SPEARMAN $\uparrow$ & OVER@100 $\uparrow$ \\
    \midrule
    \multirow{4}[4]{*}{Non-GNN based} & LR    & 0.9051±0.0137 & \textbf{0.5515±0.0110} & 0.9407±0.0076 & 0.7837±0.0044 & 0.3880±0.0232 \\
          & \textbf{  +LENIE} & \textbf{0.8952±0.0155} & 0.5528±0.0063 & \textbf{0.9464±0.0047} & \textbf{0.7928±0.0053} & \textbf{0.4080±0.0471} \\
\cmidrule{2-7}          & MLP   & 1.1125±0.0163 & \textbf{0.6926±0.0134} & 0.9246±0.0065 & 0.7075±0.0109 & \textbf{0.3320±0.0387} \\
          & \textbf{  +LENIE} & \textbf{1.0974±0.0236} & 0.7108±0.0312 & \textbf{0.9251±0.0085} & \textbf{0.7158±0.0166} & 0.3100±0.0261 \\
    \midrule
    \multirow{10}[9]{*}{GNN based} & GCN   & 2.4122±0.3064 & 1.5846±0.3339 & 0.8778±0.0147 & 0.4660±0.0288 & 0.1800±0.0352 \\
          & \textbf{  +LENIE} & \textbf{1.6832±0.1587} & \textbf{1.0645±0.1265} & \textbf{0.8886±0.0229} & \textbf{0.4787±0.0467} & \textbf{0.2060±0.0585} \\
\cmidrule{2-7}          & RGCN  & \textbf{4.0882±1.5456} & 2.5390±0.9453 & 0.8547±0.0189 & 0.2859±0.0264 & 0.1620±0.0608 \\
          & \textbf{  +LENIE} & 4.9574±1.9854 & \textbf{2.2684±1.5325} & \textbf{0.8624±0.0124} & \textbf{0.4501±0.0233} & \textbf{0.1720±0.0293} \\
\cmidrule{2-7}          & GraphSAGE & 0.8645±0.0156 & \textbf{0.5136±0.0103} & \textbf{0.9491±0.0081} & 0.8076±0.0088 & 0.4380±0.0160 \\
          & \textbf{  +LENIE} & \textbf{0.8543±0.0244} & 0.5168±0.0123 & 0.9489±0.0093 & \textbf{0.8111±0.0097} & \textbf{0.4540±0.0361} \\
\cmidrule{2-7}          & GENI  & 0.9456±0.0295 & 0.5487±0.0165 & 0.9174±0.0139 & 0.7710±0.0094 & 0.3260±0.0706 \\
          & \textbf{  +LENIE} & \textbf{0.9203±0.0304} & \textbf{0.5394±0.0299} & \textbf{0.9301±0.0148} & \textbf{0.7798±0.0249} & \textbf{0.3700±0.0576} \\
\cmidrule{2-7}          & LICAP & \textbf{0.9373±0.0578} & \textbf{0.5596±0.0431} & 0.9409±0.0086 & 0.7839±0.0130 & 0.4040±0.0546 \\
          & \textbf{  +LENIE} & 0.9549±0.0858 & 0.5800±0.0926 & \textbf{0.9437±0.0080} & \textbf{0.7853±0.0193} & \textbf{0.4160±0.0520} \\
    \midrule
          &       & \multicolumn{5}{c}{TMDB5K} \\
\cmidrule{3-7}          & Methods & RMSE $\downarrow$ & MedianAE $\downarrow$ & NDCG@100 $\uparrow$ & SPEARMAN $\uparrow$ & OVER@100 $\uparrow$ \\
    \midrule
    \multirow{4}[4]{*}{Non-GNN based} & LR    & 1.1427±0.0217 & 0.7688±0.0341 & 0.7507±0.0337 & 0.3569±0.0321 & 0.3300±0.0303 \\
          & \textbf{  +LENIE} & \textbf{0.8957±0.0261} & \textbf{0.5832±0.0157} & \textbf{0.8760±0.0237} & \textbf{0.6587±0.0284} & \textbf{0.4900±0.0434} \\
\cmidrule{2-7}          & MLP   & 1.2940±0.0180 & 0.9049±0.0268 & 0.7192±0.0270 & 0.2751±0.0100 & 0.2680±0.0397 \\
          & \textbf{  +LENIE} & \textbf{1.0674±0.0255} & \textbf{0.7035±0.0281} & \textbf{0.8386±0.0184} & \textbf{0.5539±0.0255} & \textbf{0.4320±0.0256} \\
    \midrule
    \multirow{10}[9]{*}{GNN based} & GCN   & 0.9656±0.0486 & 0.6323±0.0288 & 0.8620±0.0233 & 0.6588±0.0471 & 0.4480±0.0417 \\
          & \textbf{  +LENIE} & \textbf{0.9147±0.0246} & \textbf{0.5831±0.0180} & \textbf{0.8774±0.0252} & \textbf{0.7418±0.0183} & \textbf{0.4840±0.0578} \\
\cmidrule{2-7}          & RGCN  & \textbf{0.8703±0.0207} & 0.5926±0.0151 & \textbf{0.8656±0.0171} & 0.7282±0.0240 & 0.4640±0.0314 \\
          & \textbf{  +LENIE} & 0.8726±0.0781 & \textbf{0.5824±0.0415} & 0.8636±0.0231 & \textbf{0.7500±0.0389} & \textbf{0.4740±0.0273} \\
\cmidrule{2-7}          & GraphSAGE & 0.9897±0.0159 & 0.6629±0.0278 & 0.8297±0.0322 & 0.5467±0.0318 & 0.4160±0.0432 \\
          & \textbf{  +LENIE} & \textbf{0.8250±0.0185} & \textbf{0.5622±0.0212} & \textbf{0.8860±0.0065} & \textbf{0.7127±0.0167} & \textbf{0.5240±0.0196} \\
\cmidrule{2-7}          & GENI  & 0.7741±0.0291 & 0.5039±0.0326 & 0.8522±0.0132 & 0.7345±0.0219 & 0.4240±0.0492 \\
          & \textbf{  +LENIE} & \textbf{0.7114±0.0134} & \textbf{0.4586±0.0158} & \textbf{0.9146±0.0128} & \textbf{0.7943±0.0114} & \textbf{0.5860±0.0049} \\
\cmidrule{2-7}          & LICAP & 0.7861±0.0161 & 0.5116±0.0358 & 0.8851±0.0195 & 0.7306±0.0178 & 0.5180±0.0172 \\
          & \textbf{  +LENIE} & \textbf{0.7446±0.0241} & \textbf{0.4792±0.0364} & \textbf{0.8996±0.0142} & \textbf{0.7685±0.0062} & \textbf{0.5620±0.0376} \\
    \midrule
          &       & \multicolumn{5}{c}{MUSIC10K} \\
\cmidrule{3-7}          & Methods & RMSE $\downarrow$ & MedianAE $\downarrow$ & NDCG@100 $\uparrow$ & SPEARMAN $\uparrow$ & OVER@100 $\uparrow$ \\
    \midrule
    \multirow{4}[4]{*}{Non-GNN based} & LR    & 0.1095±0.0047 & 0.0652±0.0014 & 0.8702±0.0171 & 0.3566±0.0199 & 0.4420±0.0445 \\
          & \textbf{  +LENIE} & \textbf{0.0943±0.0026} & \textbf{0.0552±0.0011} & \textbf{0.9141±0.0173} & \textbf{0.5872±0.0105} & \textbf{0.5420±0.0366} \\
\cmidrule{2-7}          & MLP   & 0.4606±0.0342 & 0.3045±0.0196 & 0.7229±0.0244 & 0.0349±0.0523 & 0.1500±0.0237 \\
          & \textbf{  +LENIE} & \textbf{0.4528±0.0286} & \textbf{0.3040±0.0117} & \textbf{0.7270±0.0142} & \textbf{0.0775±0.0363} & \textbf{0.1560±0.0206} \\
    \midrule
    \multirow{10}[10]{*}{GNN based} & GCN   & 0.1049±0.0030 & 0.0589±0.0016 & 0.8854±0.0199 & 0.4440±0.0326 & \textbf{0.4680±0.0387} \\
          & \textbf{  +LENIE} & \textbf{0.0972±0.0039} & \textbf{0.0548±0.0022} & \textbf{0.9017±0.0119} & \textbf{0.5770±0.0237} & 0.4640±0.0350 \\
\cmidrule{2-7}          & RGCN  & 0.0972±0.0037 & 0.0565±0.0016 & 0.8974±0.0087 & 0.5377±0.0097 & 0.4880±0.0232 \\
          & \textbf{  +LENIE} & \textbf{0.0905±0.0028} & \textbf{0.0524±0.0033} & \textbf{0.9217±0.0092} & \textbf{0.6320±0.0148} & \textbf{0.5580±0.0299} \\
\cmidrule{2-7}          & GraphSAGE & 0.0983±0.0039 & 0.0579±0.0020 & 0.9076±0.0111 & 0.4591±0.0195 & 0.5240±0.0338 \\
          & \textbf{  +LENIE} & \textbf{0.0875±0.0030} & \textbf{0.0493±0.0008} & \textbf{0.9344±0.0047} & \textbf{0.6322±0.0228} & \textbf{0.5620±0.0240} \\
\cmidrule{2-7}          & GENI  & 0.0985±0.0045 & 0.0597±0.0021 & 0.8872±0.0065 & 0.4928±0.0262 & 0.4780±0.0194 \\
          & \textbf{  +LENIE} & \textbf{0.0901±0.0025} & \textbf{0.0529±0.0018} & \textbf{0.9273±0.0053} & \textbf{0.6399±0.0173} & \textbf{0.5640±0.0162} \\
\cmidrule{2-7}          & LICAP & 0.0986±0.0045 & 0.0598±0.0029 & 0.9130±0.0133 & 0.4717±0.0331 & 0.5420±0.0397 \\
          & \textbf{  +LENIE} & \textbf{0.0887±0.0043} & \textbf{0.0499±0.0026} & \textbf{0.9304±0.0084} & \textbf{0.6190±0.0174} & \textbf{0.5680±0.0331} \\
    \bottomrule
    \end{tabular}
  \label{exp_models}
\end{table*}

\subsection{Case Study: LENIE Addressing Semantic Deficiencies} \label{RQ6}

\begin{figure*}
    \centering
    \includegraphics[scale=0.42]{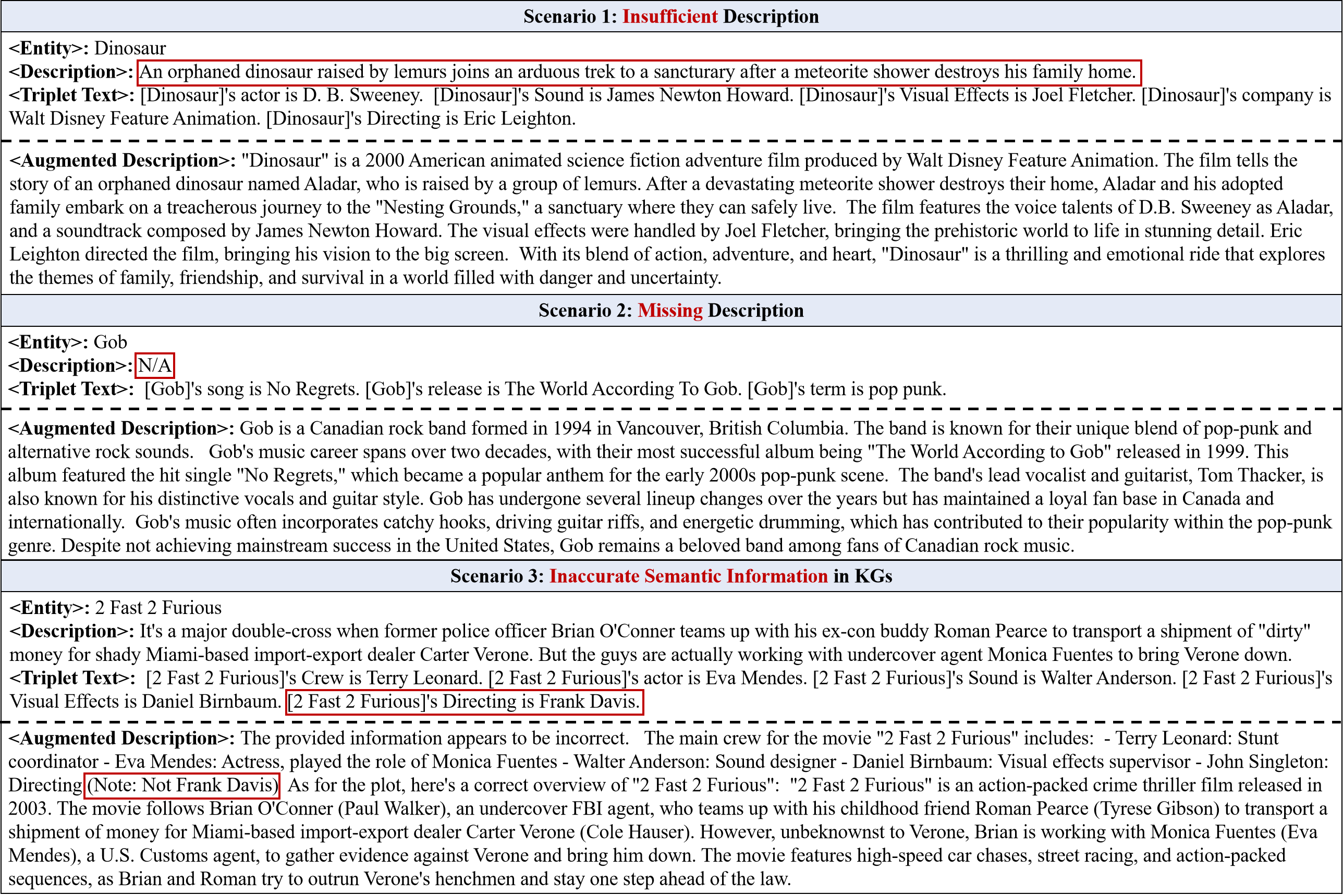}
    \caption{A case study of LENIE's semantic augmentation in three scenarios. Above the dashed line is the semantic information extracted by LENIE from KGs, and below is the augmented description generated by LLMs given the semantic information.}
    \label{Fig:text}
\end{figure*}

LENIE is capable of addressing semantic deficiencies in KGs, such as insufficient descriptions, missing descriptions, and inaccurate semantic information, as illustrated in the three scenarios shown in Figure \ref{Fig:text}. 

In Scenario 1, the movie node "Dinosaur" from the TMDB5K dataset has a description that is merely a brief plot summary, which can be considered an insufficient description. LLMs can not only integrates the node's original description and the semantic information contained in the triplet text, but also leverages its rich internal real-world knowledge to generate a more informative augmented description for the node. In Scenario 2, the singer node "Gob" from the MUSIC10K dataset lacks a description, which can be considered a missing description. LLMs can accurately understand a real-world node based solely on triplet text extracted from the KG and generate correct and enriched augmented descriptions for it. In Scenario 3, the movie node "2 Fast 2 Furious" also comes from the TMDB5K dataset. However, one of the triplet sentences incorrectly states that the directing is by Frank Davis, who is actually the assistant director, not the director. LLMs not only integrate the node's original description and information from the triplet text but also perform correctness checks, correcting inaccurate information, and generate a more accurate augmented description for the node.

Overall, LENIE leverages the complementary advantages of LLMs and KGs to provide more comprehensive and accurate descriptions for nodes, enriching the semantic information in KGs and better supporting NIE tasks on KGs.

\section{Conclusion}\label{conclusion}
The semantic deficiencies in KGs, such as insufficient descriptions, missing descriptions, and inaccurate semantic information, would restrict the performance of the existing NIE methods that utilize semantic information for NIE tasks. To address these challenges, this work proposed a novel NIE framework to leverage the advantages of LLMs for sematic augmentation and supporting downstream NIE tasks. To be specific, the proposed LENIE employed a clustering-based triplet sampling strategy to extract semantically diverse information from KGs. After that, We fed the node-specific adaptive prompts, which combine the sampled triplets and original node descriptions, into LLMs for generating the richer and more precise augmented descriptions. These augmented descriptions were then used to boost the performance of various existing NIE methods. The comprehensive experiments have demonstrated that LENIE can improve the performance of several existing NIE methods and indeed achieved the new state-of-the-art performance, and meanwhile confirmed the effectiveness of key designs in LENIE and the capability of LENIE in addressing the sematic deficiencies. 

For future work, one interesting direction is to also include the known importance scores for semantic augmentation using LLMs, as previous work \cite{zhang2024label} has shown the effectiveness of using label information in further improving NIE performance. The proposed LENIE is a flexible framework, and therefore another promising future direction is to extend the idea of semantic augmentation with LLMs and the clustering-based triplet sampling strategy for further enhancing other KG-related tasks \cite{li2024llm} and applications \cite{fang2023knowledge}.

\newpage
\bibliographystyle{IEEEtran}
\bibliography{main.bib}

\end{document}